\documentclass[letterpaper, 10 pt, conference]{ieeeconf}  

\IEEEoverridecommandlockouts                              

\usepackage{amsmath} 
\usepackage{amssymb}  
\usepackage{svg}
\usepackage{caption}
\usepackage{multirow}
\usepackage{graphics}
\usepackage{mathtools}
\usepackage{makecell}
\usepackage[linesnumbered,ruled,vlined]{algorithm2e}

\title{\LARGE \bf
Improved 3D Point-Line Mapping Regression for Camera Relocalization
}

\author{Bach-Thuan Bui$^{1}$, Huy-Hoang Bui$^{1}$, Yasuyuki Fujii$^{2}$, Dinh-Tuan Tran$^{2}$, and Joo-Ho Lee$^{2}$
\thanks{$^{1}$Graduate School of Information Science and Engineering, Ritsumeikan University, Japan.
        }%
\thanks{$^{2}$College of Information Science and Engineering, Ritsumeikan University, Japan.}%
}

\begin{document}

\maketitle
\thispagestyle{empty}
\pagestyle{empty}

\begin{abstract}
In this paper, we present a new approach for improving 3D point and line mapping regression for camera re-localization. Previous methods typically rely on feature matching (FM) with stored descriptors or use a single network to encode both points and lines. While FM-based methods perform well in large-scale environments, they become computationally expensive with a growing number of mapping points and lines. Conversely, approaches that learn to encode mapping features within a single network reduce memory footprint but are prone to overfitting, as they may capture unnecessary correlations between points and lines. We propose that these features should be learned independently, each with a distinct focus, to achieve optimal accuracy. To this end, we introduce a new architecture that learns to prioritize each feature independently before combining them for localization. Experimental results demonstrate that our approach significantly enhances the 3D map point and line regression performance for camera re-localization. The implementation of our method will be publicly available at: https://github.com/ais-lab/pl2map/.

\end{abstract}

\section{Introduction}

Visual (re)localization involves estimating the camera pose within a predefined map using visual input. It is a key component in mixed reality and robotics systems. Most current approaches rely primarily on point-based methods for localization and mapping \cite{sarlin2019coarse, brachmann2021visual, brachmann2023accelerated, shu2023structure, wang2024hscnet++}. However, integrating line-assisted features offers a more comprehensive understanding of scene structures and geometric details, enhancing the adaptability and effectiveness of visual re-localization in various applications \cite{gao2022pose, liu20233d, xu2023airvo}.

Methods utilizing both point and line features for visual localization can be broadly classified into feature matching (FM) and regression-based approaches. Both rely on pre-built maps generated by SLAM or SfM techniques \cite{xu2023airvo, liu20233d, zhang2023pl, gomez2019pl, shu2023structure}. FM-based methods \cite{gao2022pose, liu20233d} offer high accuracy and scalability but require resource-intensive components, such as image databases, retrieval systems, and feature matchers for points and lines. In contrast, Bui et al. \cite{bui2024representing} introduced the first regression-based attempt for jointly encoding 3D point and line coordinates within a single neural network. This approach offers advantages like reduced storage, faster inference, and enhanced privacy through implicit feature representation. However, it still suffers from challenges, particularly the imbalance between the number of points and lines, which introduces bias during training and degrades accuracy by overemphasizing one feature type over the other. Additionally, both \cite{bui2024representing} and previous points-based regression approaches \cite{brachmann2021visual, bui2024d2s, wang2024hscnet++, brachmann2023accelerated} struggle with learning from noisy or poor-quality features, which can adversely influence training and lead to suboptimal performance \cite{nguyen2024focustune}.

\begin{figure}
    \centering
    \includegraphics[width=\linewidth]{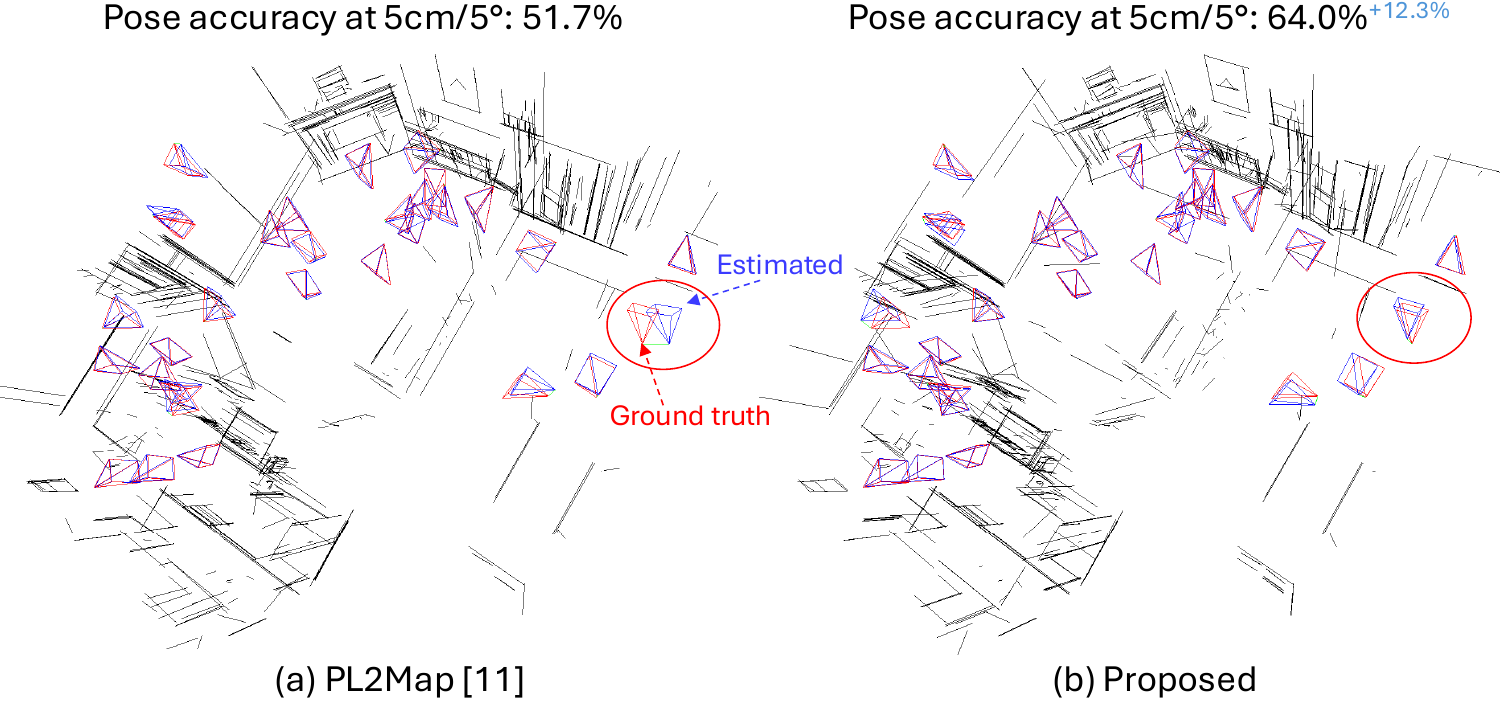}
    \caption{Camera localization results and predicted line maps in Scene-1, Indoor-6 \cite{do2022learning} by PL2Map \cite{bui2024representing} (left) and proposed method (right). The proposed method gives accurate camera pose estimates by addressing problems of imbalance and noisy features in joint training.}
    \label{fig1}
\end{figure}

To address the aforementioned issues, we treat line features as complementary to the sparse point-based visual localization process. We redesign the network architecture to incorporate separate streams for points and lines, allowing the model to learn each feature independently. This approach mitigates the risk of bias, overfitting, or performance degradation due to the imbalance between points and lines. Similar to \cite{bui2024representing}, we treat points and lines as two distinct sets of unordered descriptors, but restructure the network into two separate regression branches. For both streams, we introduce an early pruning layer to filter out unimportant features before applying regression. Since sparse features lack natural connectivity, we use multiple self-attention layers to enhance their robustness before regressing 3D coordinates via a multi-layer perceptron (MLP).  In contrast to the point branch, for the line regression stream, we adapt a line transformer encoder and use a single self-attention layer before pruning, given the relatively small number of lines per image. We then apply a smaller MLP to regress the 3D coordinates of the selected lines. In summary, our contributions are as follows: 

\begin{itemize}
\item We present an enhanced 3D point-line regression method that effectively addresses the imbalance between points and lines in visual localization mapping.
\item We introduce a redesigned regression architecture with an early learnable pruning layer, allowing the network to focus on critical features.
\item We validate our method on two datasets, including those with challenging and dynamic conditions, showing a significant performance improvement over previous regression-based methods.
\end{itemize}

Fig. \ref{fig1} shows a sub-sample of camera localization results by the proposed method, compared to the prior work on joint training points and lines.

\section{Related Work}

\subsection{Visual Systems with Line-assisted}

Recent visual SLAM, visual odometry (VO), and SfM methods \cite{gomez2016pl, pumarola2017pl, gomez2019pl, yang2019visual, liu20233d} have demonstrated significant performance improvements by incorporating both point and line features. Early systems relied on hand-crafted line features, such as gradient grouping techniques \cite{akinlar2011edlines, von2008lsd}. However, with the rise of deep learning, learning-based line descriptors have been developed to enhance robustness and repeatability \cite{pautrat2021sold2, yoon2021line, pautrat2023deeplsd}. These advancements have further optimized the above systems, improving overall performance in a variety of settings \cite{gao2022pose, liu20233d, xu2023airvo}.

\subsection{FM-based Visual Localization}
Early visual localization relied on efficient image retrieval, where the environment was represented by a database of images with known poses \cite{chen2011city, arandjelovic2015dislocation, arandjelovic2016netvlad}. The goal was to find the most similar images in the database based on a query image. Although these methods lacked accuracy due to database sampling limitations, they served as a useful initialization for pose refinement. FM-based methods improved upon this by matching local features between query and database images to recover the camera pose. Hloc \cite{sarlin2019coarse, sarlin2020superglue} emerged as a comprehensive visual localization system, incorporating advanced feature detection \cite{detone2018superpoint,revaud2019r2d2, dusmanu2019d2,potje2024xfeat} and matching techniques \cite{sarlin2020superglue, lindenberger2023lightglue, sun2021loftr, giang2023topicfm}. Limap \cite{liu20233d} extended these capabilities by integrating line features, leveraging deep line detection \cite{pautrat2021sold2,yoon2021line, pautrat2023deeplsd} and matching methods \cite{pautrat2023gluestick, yoon2021line, pautrat2021sold2}, significantly improving localization accuracy. However, the need to store large amounts of descriptors for both points and lines raises concerns regarding storage overhead and privacy. As the environment size increases, the computational costs associated with these methods become prohibitive, particularly in real-time applications \cite{do2024improved}.

\subsection{Learning Surrogate Maps}
One potential solution to the above-mentioned problem is the use of regression-based approaches \cite{brachmann2021visual, brachmann2023accelerated, bui2024d2s, bui2024leveraging}. These methods learn to directly establish 2D-3D correspondences between 2D pixels and 3D coordinates, eliminating the need for feature matching and database retrieval steps. Early work in this area utilized random forests \cite{brachmann2016uncertainty, cavallari2017fly, cavallari2019real}, while more recent approaches have adopted convolutional neural networks (CNNs) \cite{brachmann2021visual, brachmann2023accelerated}. By embedding the environmental map into the network weights, these methods significantly improve inference speed and reduce storage requirements. To further enhance this category of algorithms, PL2Map \cite{bui2024representing} was introduced to jointly regress both line and point features, resulting in a more robust map representation. However, this approach still does not fully address the challenge of multi-feature regression, leading to suboptimal performance during network training. 


\section{Proposed Method}
\begin{figure*}
    \centering
    \includegraphics[width=1.03\linewidth]{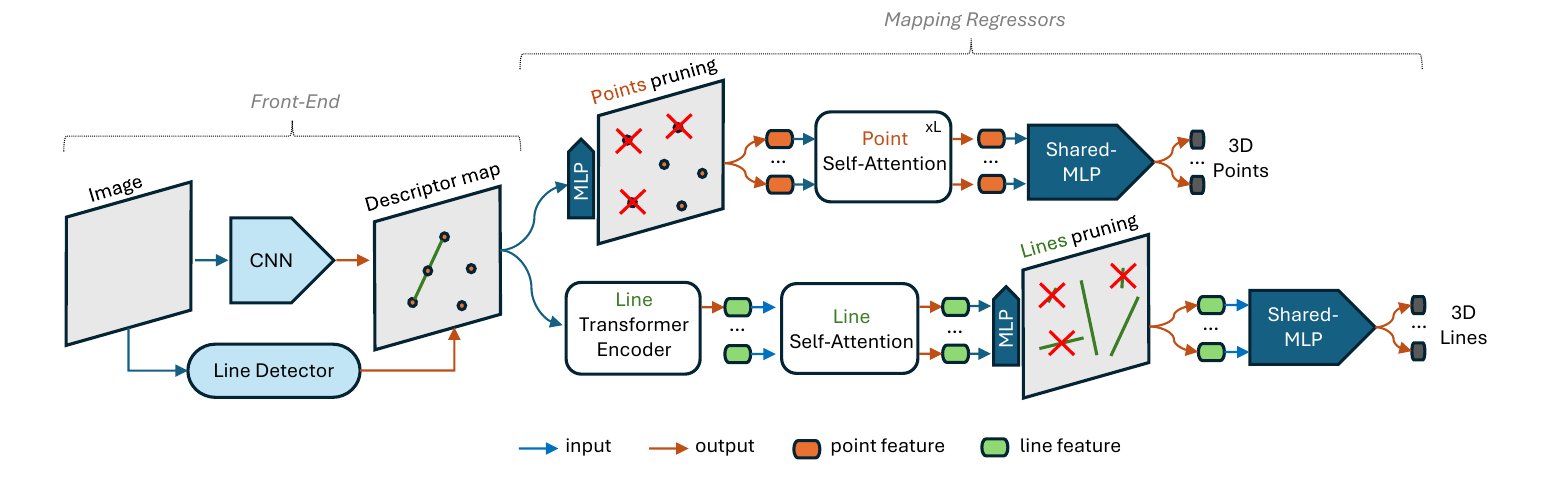}
    \caption{\textbf{Proposed architecture.} The architecture focuses on the distinct regression of 3D points and lines, consisting of two main components: (1) the Front-End, which preprocesses and jointly extracts point and line descriptors via a shared feature extractor, and (2) the Mapping Regressors, which include separate regression branches dedicated to point and line maps.}
    \label{network_pipeline}
\end{figure*}
We now present the proposed method in detail, including the problem statement and the comprehensive idea for enhancing 3D point and line regression. 
\subsection{Problem Statement}
Given a point-based SfM model $\mathcal{S}^{p} \leftarrow \{\mathbf{P}_k \in \mathbb{R}^{3} \mid k = 1, 2, \dots, M\}$ and a line-based SfM model $\mathcal{S}^{l} \leftarrow  \{\mathbf{L}_v \in \mathbb{R}^{6} \mid v = 1, 2, \dots, N\} $, created using a same set of reference images $ \{\mathbf{I}_{i}\}^{n}$. Let $\mathcal{P}_i^{p} = \{\mathbf{d}_{i,j}^{p} \leftarrow \{k, \text{None}\} \mid j = 1, 2, \dots, M_{i}\}$ represent the set of point descriptors extracted in image $\mathbf{I}_i$, where $k$ is the index of the corresponding 3D point $\mathbf{P}_k$. Similarly,  let $\mathcal{P}_i^{l} = \{\mathbf{d}_{i,j}^{l}  \leftarrow \{v, \text{None}\}\} \mid j = 1, 2, \dots, N_{i}\}$ represent the set of line descriptors from the same image. Due to the significant imbalance between $M$ and $N$, and the presence of noisy 2D points and lines, which are not consistently reliable for learning accurate 3D coordinates, we propose an adaptive regressor $\mathcal{F}(.)$ that takes $\mathcal{P}_i^{p}$ and $\mathcal{P}_i^{l}$ as inputs and selectively learns to output robust 3D coordinates. The ultimate objective is to estimate the six degrees of freedom (6 DOF) camera pose $\mathbf{T} \in \mathbb{R}^{4\times4}$ for any new query image $\mathbf{I}$ from the same environment.

\subsection{Point-Lines Regression in Focus Mode}
In the following sections, we present a detailed description of the proposed framework, which can selectively learn to produce 2D-3D correspondences for both points and lines.

\subsubsection{Front-End}

Similar to previous works \cite{bui2024d2s, bui2024representing, brachmann2023accelerated}, the proposed method employs a pre-trained feature extractor to obtain descriptors for keypoints and keylines. We use SuperPoint \cite{detone2018superpoint} for keypoints and their descriptors, while also leveraging its ability to represent line features \cite{von2008lsd, pautrat2023deeplsd}. This approach incurs a low cost, as it eliminates the need for an additional line descriptor extractor for subsequent regression.

Unlike the prior work, \cite{bui2024representing}, which attempted to encapsulate both point and line features within the same network, we argue that each feature may have distinct properties that require focused learning. The inherent imbalance between these features can also impose significant challenges during training, potentially leading to suboptimal performance for both point and line regression. Therefore, we propose separate regression streams for each feature, as detailed in the following sections. We show complete learning architecture in Fig. \ref{network_pipeline}.

\subsubsection{Point Regressor Branch}

The extracted descriptors from the previous step are filtered through a simple pruning layer. Specifically, the pruning probability for each descriptor is computed as follows: 

\begin{equation}
    \alpha_{j}^{p} = \text{Sigmoid}\bigl(\phi^{p}(\mathbf{d}_{j}^{p}) \bigl) \in [0,1],
    \label{sigmoid_point}
\end{equation}
where $\phi^{p}$ is an MLP shared parameter for all point descriptors. 

We then use a threshold $\delta^{p}$ to prune descriptors with $\alpha_{j}^{p} \le \delta^{p}$ and keep only the robust ones with $\alpha_{j}^{p} > \delta^{p}$ for later refinement using a multi-graph attention network. 

Inspired by the success of previous works \cite{sarlin2020superglue, lindenberger2023lightglue, bui2024d2s, bui2024representing}, we refine the reliable descriptors using a multi-self-attention network, where each descriptor is updated as follows: 

\begin{equation}
\begin{aligned}
\prescript{(m+1)}{}{\mathbf{d}_{j}} = \prescript{(m)}{}{\mathbf{d}_{j}} + \phi_{l}\bigg(\bigg[\prescript{(m)}{}{\mathbf{d}_{j}}|| a_{m}\big(\prescript{(m)}{}{\mathbf{d}_{j}}, \mathcal{E}\big) \bigg]\bigg),
\label{attention_equation}
\end{aligned}
\end{equation}

where $\prescript{(l)}{}{\mathbf{d}_{j}}$ is the intermediate descriptor for element $j$ in  layer $m$, $\mathcal{E}$ is reliable descriptors set in same layer $m$, and $a_{m}(.)$ is the self-attention from \cite{vaswani2017attention} applied to the same set of descriptors, which is calculated as: 

\begin{equation}
\begin{aligned}
a_{l}(\prescript{(m)}{}{\mathbf{d}_{j}}, \mathcal{E}) = \sum_{j:(i,j)\in\mathcal{E}}{\text{Softmax}_{j}\big(\mathbf{q}_{j}^{T}\mathbf{k}_{i}/\sqrt{D}\big)\mathbf{v}_{i}},
\label{ori_equation}
\end{aligned}
\end{equation}

where $\mathbf{q}_{j}$, $\mathbf{k}_{i}$, $\mathbf{v}_{i}$ are the linear projections of descriptors $\mathbf{d}$. 

Finally, we use a shared MLP to linearly map descriptors to 3D coordinates: 
\begin{equation}
\begin{aligned}
\hat{\mathbf{P}}_{j}= \text{MLP}\big( \prescript{(m)}{}{\mathbf{d}^{p}_{j}} \big) \in \mathbb{R}^{3}
\label{ori_equation}
\end{aligned}
\end{equation}

\subsubsection{Line Regressor Branch}
Since the line branch shares the feature extractor with keypoints, where the extracted descriptors are point-based, we adapt the same line descriptor encoder proposed in \cite{bui2024representing} to obtain a coarse set of low-dimensional descriptors. Additionally, we observe that classifying robust lines at this early stage is challenging, as line descriptors have not yet been established and the number of lines is small. Therefore, in this line regressor branch, we design the pruning layers to be applied at a later stage.  

\begin{figure}
    \centering
    \includegraphics[width=0.85\linewidth]{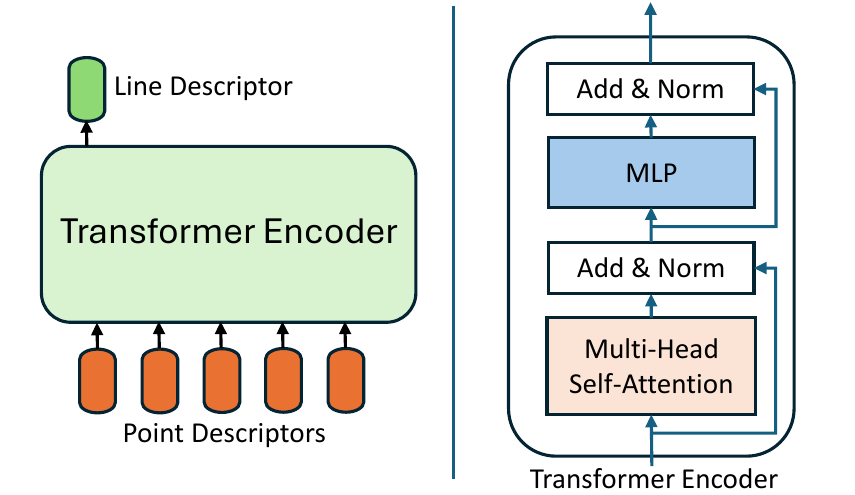}
    \caption{\textbf{Line transformer encoder.} We use a transformer-based model to encode a sample of $C$ point descriptors to a single line descriptor, left figure. On the right, we illustrate the transformer architecture in detail.}
    \label{line_transformer}
\end{figure}

For a line segment $\mathbf{l}_{j} \in \mathbb{R}^{4}$, we apply a single transformer model to encode $C$ descriptors, sampled from the $\mathbf{l}_{j}$ segment, into a unique descriptor $\mathbf{d}^{l}_{j} \in \mathbb{R}^{256}$. In particular, we adopt a simplified version of the transformer encoder, skipping the use of position encoders. This process is illustrated in Fig. \ref{line_transformer}. The self-attention in the transformer encoder is implemented as described in Eq. \ref{attention_equation}.

In contrast to the point branch, we employ only a single self-attention layer to refine the line descriptors from their coarse representation. This is due to the relatively smaller number of lines compared to points, and their occasional absence in certain regions. The implementation follows directly from Eq. \ref{attention_equation}, where the resulting fine line descriptor is denoted as $\prescript{(1)}{}{\mathbf{d}_{j}^{l}}$.

\begin{table*}
\centering
\caption{\textbf{Results on 7Scenes} \cite{shotton2013scene}. We report the results in median position error (cm), rotation error (degree), and accuracy with errors lower than 5cm/5$^{\circ}$ when using both point and line features. The results highlighted in \textcolor{blue}{blue} color are the best.}
\begin{tabular}{c|c|c|c|c|c} 
\hline
\multicolumn{6}{c}{7Scenes}                                                                                                                                                                                         \\
\hline
           & Hloc $^\text{point}$ \cite{sarlin2019coarse, sarlin2020superglue}              & PtLine $^\text{point+line}$ \cite{gao2022pose} & Limap $^\text{point+line}$ \cite{liu20233d}        & Pl2Map $^\text{point+line}$ \cite{bui2024representing}       & \textbf{Proposed} $^\text{point+line}$         \\ 
           & (cm / deg. / \%)   & (cm / deg. / \%) & (cm / deg. / \%)   & (cm / deg. / \%)   & (cm / deg. / \%)                     \\
\hline
Chess                & 2.4 / 0.84 / 93.0                  & 2.4 / 0.85 / 92.7                & 2.5 / 0.85 / 92.3                  & \textcolor{blue}{1.9} / 0.63 / 96.0 & 2.0 / \textcolor{blue}{0.62 / 96.1}                   \\
Fire                 & 2.3 / 0.89 / 88.9                  & 2.3 / 0.91 / 87.9                & 2.1 / 0.84 / \textcolor{blue}{95.5} & \textcolor{blue}{1.9 /} 0.80 / 94.0 & \textcolor{blue}{1.9 / 0.77} / 91.8                   \\
Heads                & \textcolor{blue}{1.1} / 0.75 / 95.9 & 1.2 / 0.81 / 95.2                & \textcolor{blue}{1.1 }/ 0.76 / 95.9 & \textcolor{blue}{1.1 / 0.71 }/ 98.2 & \textcolor{blue}{1.1 / 0.70 / 98.8}                   \\
Office               & 3.1 / 0.91 / 77.0                  & 3.2 / 0.96 / 74.5                & 3.0 / 0.89 / 78.4                  & 2.7 / 0.74 / 84.3                  & \textcolor{blue}{2.5 / 0.69 / 87.2}                   \\
Pumpkin              & 5.0 / 1.32 / 50.4                  & 5.1 / 1.35 / 49.0                & 4.7 / 1.23 / 52.9                  & 3.4 / 0.93 / 64.1                  & \textcolor{blue}{2.9 / 0.86 / 68.3}                   \\
RedKitchen           & 4.2 / 1.39 / 58.9                  & 4.3 / 1.42 / 58.0                & 4.1 / 1.39 / 60.2                  & \textcolor{blue}{3.7 / 1.10 }/ 68.9 & \textcolor{blue}{3.7} / 1.12 / \textcolor{blue}{71.0}  \\
Stairs               & 5.2 / 1.46 / 46.8                  & 4.8 / 1.33 / 51.9                & \textcolor{blue}{3.7 / 1.02 / 71.1} & 7.6 / 2.0 / 33.3                   & 7.2 / 1.92 /  34.0                                   \\
\hline
\end{tabular}
\label{7scenes_results}
\end{table*}

Next, we apply a simple pruning layer to remove lines that are either unimportant or have low reliability for the final regression, calculated as follows:

\begin{equation}
    \alpha_{j}^{l} = \text{Sigmoid}\bigl(\phi^{l}(\prescript{(1)}{}{\mathbf{d}_{j}^{l}})\bigl) \in [0,1],
\end{equation}

where $\alpha_{j}^{l}$ is the reliability probability for line $j$ and $\phi^{l}$ is a MLP. In here, also keep only the line descriptors with $\alpha_{j}^{l} > \delta^{l}$, where $\delta^{l}$ is the prunning threshold for lines. 

Finally, we linearly map the line descriptors to their corresponding 3D segment coordinates as follows:

\begin{equation}
\begin{aligned}
\hat{\mathbf{L}}_{j}= \text{MLP}\big( \prescript{(1)}{}{\mathbf{d}^{l}_{j}} \big) \in \mathbb{R}^{6}.
\label{ori_equation}
\end{aligned}
\end{equation}
\subsection{Loss Function}

The predicted 3D point $\hat{\mathbf{P}_{j}}$ and line $\hat{\mathbf{L}}_{j}$ are used to optimize the model using their ground truths $\mathbf{P}_{j}$ and $\mathbf{L}_{j}$ from SfM models, calculated for each image as follows:

\begin{equation}
    \mathcal{L}_{m} = \sum_{j=1} \lVert \mathbf{P}_{j}-\hat{\mathbf{P}}_{j} \lVert_{2} + \sum_{j=1} \lVert \mathbf{L}_{j}-\hat{\mathbf{L}}_{j} \lVert_{2}.
\label{loss_m}
\end{equation}


We simultaneously optimize the pruning probability prediction using binary cross entropy (BCE) loss for both points and lines as:

\begin{equation}
\mathcal{L}_{\text{BCE}} = \sum_{j=1}^{M} \mathcal{L}_{\text{BCE}}^{p}(\hat{\alpha}_{j}^{p}, \alpha_{j}^{p}) + \sum_{j=1}^{N} \mathcal{L}_{\text{BCE}}^{l} (\hat{\alpha}^{l}_{j}, \alpha^{l}_{j}).
\end{equation}

The model is then further optimized by utilizing the available camera poses to reproject the predicted 3D points and lines onto the image plane as:

\begin{equation}
\begin{aligned}
    \mathcal{L}_{\pi} = & \sum_{j=1} \big\lVert \pi(\mathbf{T},\hat{\mathbf{P}}_{j})-\mathbf{u}_{j}^{p}\big\lVert_{2} \\ & + \sum_{j=1} \psi\big(\pi(\mathbf{T},\hat{\mathbf{L}}_{j}), \mathbf{u}_{j}^{l}\big),
\label{origin_reproject_loss}
\end{aligned}
\end{equation} 

where $\mathbf{T}$ is the ground truth pose, $\pi(.)$ is the reprojection function, $\mathbf{u}_{i}^{p} \in \mathbb{R}^{2}$ and $\mathbf{u}_{i}^{l} \in \mathbb{R}^{4}$ are the 2D positions of the point and line endpoints on the image, $\psi(.)$ is a function that calculates the distance between reprojected 3D line endspoints and its ground truth line coordinates $\mathbf{u}_{i}^{l}$.

However, the reprojection loss in Eq. \ref{origin_reproject_loss} is highly non-convex and difficult to optimize early in training. To mitigate this, we incorporate the robust projection error from \cite{brachmann2023accelerated} as follows:

\begin{equation}
\begin{aligned}
    \mathcal{L}_{\pi}^{robust} = 
\begin{cases} 
0, & \text{if } t < \theta \\
\tau(t) \tanh\bigg(\frac{\mathcal{L}_{\pi}}{\tau(t)}\bigg), & \text{otherwise},
\end{cases}
\label{reproject_loss}
\end{aligned}
\end{equation}

\begin{figure}[ht]
    \centering
    \includegraphics[width=\linewidth]{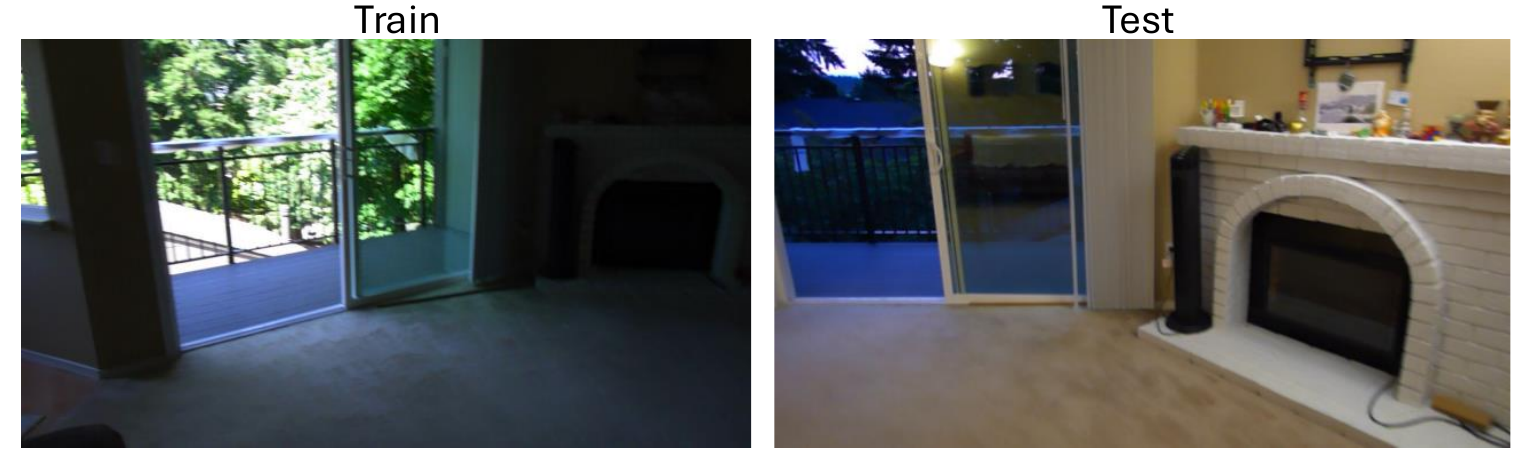}
    \caption{\textbf{Indoor-6 train-test images}. We show an example of training and testing images from the Indoor-6 dataset \cite{do2022learning}, where variations in capture times present a challenge for regression-based methods.}
    \label{indoor6_example}
\end{figure}

\begin{table*}[ht]
\centering
\caption{\textbf{Results on Indoor-6 \cite{do2022learning}.} We report the median localization errors in cm for the position,  degree ($^{\circ}$) for the orientation, and recall at 5cm/5$^{\circ}$ on the Indoor-6 dataset. The first methods row is FM-based and the second one lists regression-based methods. The best regression-based results are in \textcolor{blue}{blue} color while the best overall results are highlighted in \textbf{in bold}. }

\label{indoor6_results}
\resizebox{0.95\linewidth}{!}{%
\begin{tabular}{c|c|c|c|c|c|c} 
\hline
\multicolumn{7}{c}{Indoor-6~~}                                                                                                                                                                                                                           \\ 
\hline
\multirow{2}{*}{Method} & Scene-1                             & Scene-2a                            & Scene-3                             & Scene-4a                             & Scene-5                             & Scene-6                              \\
                        & (cm / deg. / \%)                   & (cm / deg. / \%)                       & (cm / deg. / \%)                       & (cm / deg. / \%)                        & (cm / deg. /\%)                       & (cm / deg. / \%)                        \\ 
\hline
Hloc $^{\text{point}}$ \cite{sarlin2019coarse, sarlin2020superglue}                   & 1.2 / 0.22 / 88.2                  & 1.1 /~\textbf{0.12~}/ 94.2         & 0.7 /~\textbf{0.15~}/ 95.2         & \textbf{1.0 / 0.24~}/ 94.3          & 2.0 / 0.33 /~\textbf{81.4}         & 0.6 /~\textbf{0.13~}/ 97.2          \\
Limap $^{\text{point+line}}$ \cite{liu20233d}                  & \textbf{1.1 / 0.21 / 90.7}         & \textbf{1.0 / 0.12 / 95.3}         & \textbf{0.6 / 0.15 / 96.2}         & 1.0 / 0.25 /~\textbf{95.6}          & \textbf{1.8 / 0.31 / 81.4}         & \textbf{0.5 / 0.13 / 96.9}          \\ 
\hline
PoseNet $^{\text{APR}}$ \cite{kendall2015posenet}                 & 159.0 / 7.46 / 0.0                 & - / - / -                          & 141.0 / 9.26 / 0.0                 & - / - / -                           & 179.3 / 9.37 / 0.0                 & 118.2 / 9.26 / 0.0                  \\
DSAC* $^{\text{point}}$ \cite{brachmann2021visual}                  & 12.3 / 2.06 / 18.7                 & 7.9 / 0.9 / 28.0                   & 13.1 / 2.34 / 19.7                 & 3.7 / 0.95 / 60.8                   & 40.7 / 6.72 / 10.6                 & 6.0 / 1.40 / 44.3                   \\
ACE $^{\text{point}}$ \cite{brachmann2023accelerated}                    & 13.6 / 2.1 / 24.9                  & 6.8 / 0.7 / 31.9                   & 8.1 / 1.3 / 33.0                   & 4.8 / 0.9 / 55.7                    & 14.7 / 2.3 / 17.9                  & 6.1 / 1.1 / 45.5                    \\
NBE+SLD(E) $^{\text{point}}$ \cite{do2022learning}             & 7.5 / 1.15 / 28.4                  & 7.3 / 0.7 / 30.4                   & 6.2 / 1.28 / 43.5                  & 4.6 / 1.01 / 54.4                   & 6.3 / 0.96 / 37.5                  & 5.8 / 1.3 / 44.6                    \\
NBE+SLD $^{\text{point}}$ \cite{do2022learning}                & 6.5 / 0.90 / 38.4                  & 7.2 / 0.68 / 32.7                  & 4.4 / 0.91 / 53.0                  & 3.8 / 0.94 / 66.5                   & 6.0 / 0.91 / 40.0                  & 5.0 / 0.99 / 50.5                   \\
D2S $^{\text{point}}$ \cite{bui2024d2s}                    & 4.8 / 0.81 / 51.8                  & 4.0 / 0.41 / 61.1                  & 3.6 / 0.69 / 60.0                   & 2.1 / 0.48 / 84.8                    & \textcolor{blue}{5.8 / 0.90 / 45.5} & 2.4 / 0.48 / \textcolor{blue}{75.2}  \\
PL2Map $^{\text{point+line}}$ \cite{bui2024representing}                 & 4.7 / 0.84 / 51.7                  & 4.8 / 0.49 / 53.7                  & 5.3 / 1.05 / 49.2                  & 2.0 / 0.49 / 82.9                   & 7.7 / 1.21 / 40.3                 & 3.41 / 0.61 / 64.1                  \\
\textbf{Proposed} $^{\text{point+line}}$         & \textcolor{blue}{3.6 / 0.59 / 64.0} & \textcolor{blue}{3.1 / 0.32 / 75.5} & \textcolor{blue}{3.1 / 0.62 / 65.8} & \textcolor{blue}{1.6 / 0.38 / 85.4} & 6.0 / \textcolor{blue}{0.87} / 42.9 & \textcolor{blue}{2.3 / 0.40} / 74.9  \\
\hline
\end{tabular}}
\end{table*}

where $\theta$ is the threshold used to prevent reprojection loss at the early stage, $\tau(.)$ is the threshold used to dynamically rescale $tanh(.)$. $\tau(.)$ is then calculated  as:

\begin{equation}
\begin{aligned}
    \tau(t) = \omega(t) \tau_{max} + \tau_{min},  \text{ with } \omega(t) = \sqrt{1-t^{2}}, 
\label{reproject_loss}
\end{aligned}
\end{equation} 

where $t\in(0,1)$ denotes the relative training progress. This forces the threshold $\tau$ to have a circular schedule that remains close $\tau_{max}$ at the beginning and reaches $\tau_{min}$ at the end of the training. 

Finally, we integrate all loss functions to optimize the surrogate model as follows: 

\begin{equation}
    \mathcal{L}= \delta_{m}\mathcal{L}_{m} + \delta_{\text{BCE}}\mathcal{L}_{\text{BCE}} + \delta_{\pi}\mathcal{L}_{\pi}^{robust},
    \label{final_loss}
\end{equation}

where $\delta$ is the hyperparameter coefficient used to balance three loss functions. 

\begin{figure*}
    \centering
    \includegraphics[width=0.8\linewidth]{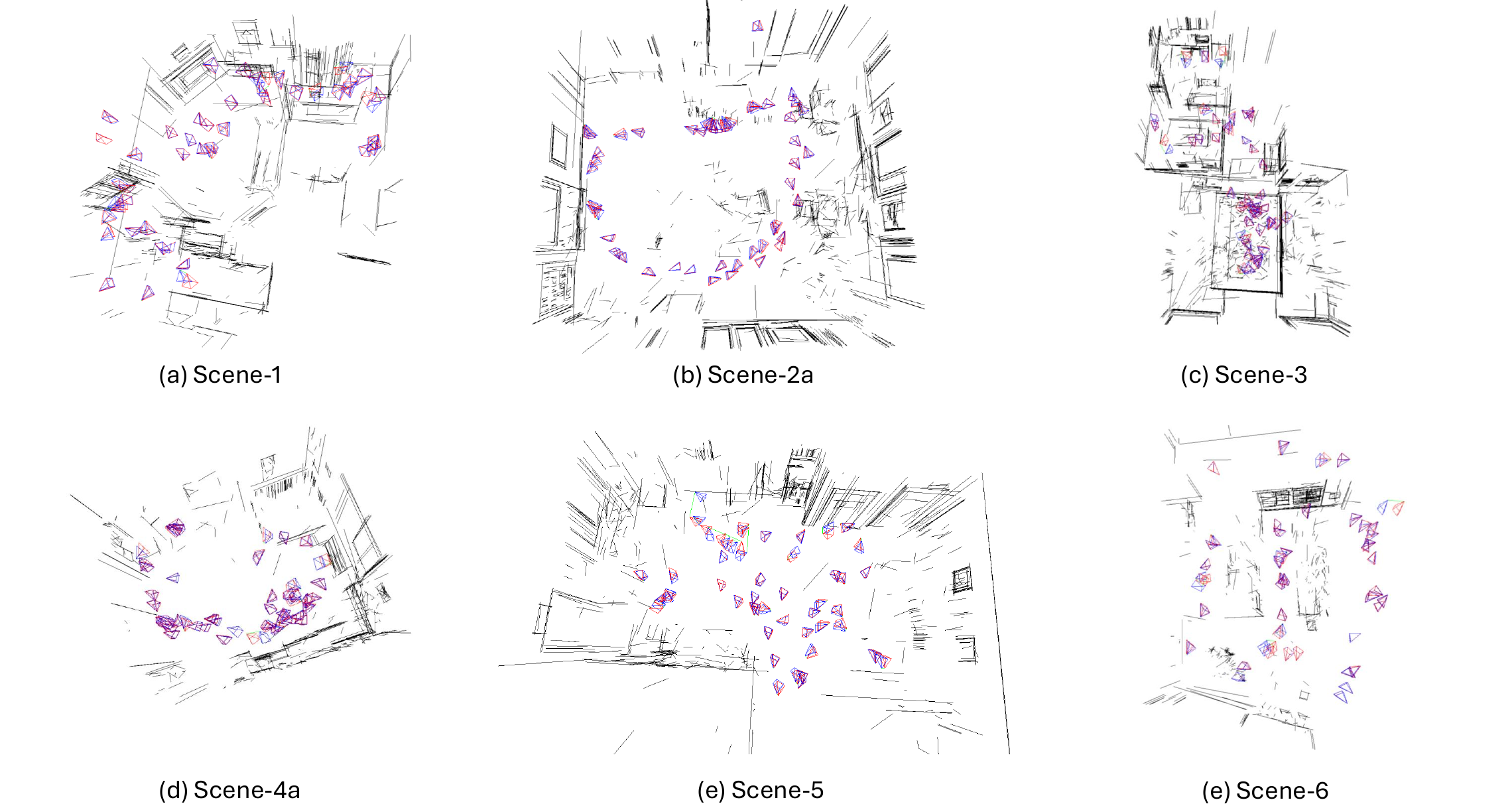}
    \caption{\textbf{Qualitative results on Indoor-6.} We display a random sample of 50 test images with their \textcolor{blue}{estimated} poses by the proposed method when using both predicted 3D points and lines. The ground truth poses are indicated in  \textcolor{red}{red} color. We additionally show the predicted 3D lines in the background using those images.}
    \label{qualitative_result_indoor6}
\end{figure*}

\section{Experiments}

We implemented our approach using PyTorch \cite{paszke2019pytorch}. The network configuration for the point branch consists of a pruning layer with $\phi^{p} = \text{MLP (256, 256, 1)}$, 5 self-attention layers, and a final regressor MLP (512, 1024, 1024, 512, 3). For the line branch, we used $C=10$ to sample the point descriptors along the line segment, followed by applying the transformer model to the sampled points using the same settings with one self-attention layer. Finally, the MLP regressor for the line branch was configured as MLP (512, 1024, 512, 6).

For the hyperparameters, we found that equal balancing of $ \delta_{m} = \delta_{\text{BCE}} = \delta_{\pi} = 1 $ provided stable training. The projection loss function was applied starting at $\theta = 0.05$. Our method was optimized using the Adam optimizer \cite{kingma2014adam}, with an initial learning rate of $2 \times 10^{-4}$, which was reduced seven times by a factor of 0.5. We trained the model for 2.5 million iterations per scene. Data augmentation was applied in all experiments, with brightness and contrast randomly adjusted by $\pm 15\%$ and $\pm 10\%$ of the input image, respectively. 

For the ground truth of 3D points, we used Hloc~\cite{sarlin2019coarse} to triangulate 2D point from correspondences and given poses, while 3D lines were similarly generated using Limap~\cite{liu20233d}. These ground-truth points and lines were also served to supervise the pruning layers in our proposed method. Specifically, we took only the most reliable 3D points and lines, \emph{i.e.}, those successfully triangulated by Hloc and Limap, as robust features, whereas failed triangulations are labeled as unreliable. During both training and evaluation, our method employed DeepLSD~\cite{pautrat2023deeplsd} for line detection and SuperPoint~\cite{detone2018superpoint} for point detection by default.

After training the models for each scene, we evaluated the re-localization accuracy of the proposed method. Specifically, we used PoseLib \cite{PoseLib} to estimate the 6DoF camera poses based on both the predicted 3D points and lines. We set the pruning threshold for points to $\delta^{p} = 0.8$ and for lines to $\delta^{l} = 0.01$. 
Notice that $\delta^{l}$ is much smaller than $\delta^{p}$, mainly because there are generally fewer lines 
than points, resulting in less training data for lines.
In our experiments, a smaller threshold for lines worked significantly better; if $\delta^{l}$ was 
increased to match $\delta^{p}$, most lines ended up being pruned, which led to a drop in performance.

\subsection{Localization Results on 7Scenes}
We evaluate the proposed approach using the 7-Scenes dataset \cite{shotton2013scene}, a small-scale environment rich in both point and line textures. The dataset comprises seven environments, each containing several thousand training and testing images. Ground truth camera poses are provided through KinectFusion \cite{izadi2011kinectfusion}. 7Scenes provides depth images for evaluation, but we use only RGB images for SfM model creation and evaluation.

For comparison with previous localization methods utilizing point and line features, we report results from PtLine \cite{gao2022pose}, Limap \cite{liu20233d}, and PL2Map \cite{bui2024representing}. PtLine and Limap are feature-matching (FM)-based methods, while PL2Map is the first point-line regression-based method. We also compare with point-only methods, such as HLoc \cite{sarlin2019coarse, sarlin2020superglue}, as a reference.

We present the results in Table \ref{7scenes_results}. The findings indicate that our proposed focus on point and line features improves localization performance across all seven scenes compared to the previous method, PL2Map \cite{bui2024representing}. Notably, the proposed method achieves accuracy gains of 3\%, 4\%, and 2\% in the Office, Pumpkin, and Red Kitchen scenes, respectively (measured at 5cm/5$^{\circ}$). Minor improvements are also observed in the Chess, Fire, Heads, and Stairs scenes compared to the previous regression-based method. However, our method still suffers from scenes with highly repetitive structures, such as the Stairs scene, similar to the limitations observed in PL2Map \cite{bui2024representing}. It is important to note that our method utilizes 3D lines from Limap as ground truth, yet it still achieves superior performance compared to Limap in these small-scale environments.

Overall, the results support our hypothesis that each learning branch should focus on representing its respective features points or lines. However, since 7Scenes is a small-scale dataset with static captured images, it is insufficient to fully validate the improvement of the proposed regression architecture. In the next section, we evaluate our method on a larger and more challenging dataset.
\subsection{Localization Results on Indoor-6}

\begin{table}
\centering
\caption{\textbf{Localization using only points in Indoor-6.} We report the median errors and accuracy of the proposed method when using only the point branch. The best results are in \textcolor{blue}{blue}.}
\begin{tabular}{c|c|c} 
\hline
\multicolumn{3}{c}{Indoor-6~~}                                     \\ 
\hline
        & ACE $^{\text{point}}$ \cite{brachmann2023accelerated}               & \textbf{Proposed} $^{\text{point}}$                             \\
        & (cm / deg. / \%)  & (cm / deg. / \%)                     \\ 
\hline
Scene-1  & 13.6 / 2.1 / 24.9 & \textcolor{blue}{4.5 / 0.80 / 54.2}  \\
Scene-2a & 6.8 / 0.7 / 31.9  & \textcolor{blue}{3.7 / 0.41 / 63.8}  \\
Scene-3  & 8.1 / 1.3 / 33.0  & \textcolor{blue}{3.4 / 0.63 / 63.2}   \\
Scene-4a & 4.8 / 0.9 / 55.7  & \textcolor{blue}{2.0 / 0.48 / 79.1}  \\
Scene-5  & 14.7 / 2.3 / 17.9 & \textcolor{blue}{8.1 / 1.32 / 38.2}  \\
Scene-6  & 6.1 / 1.1 / 45.5  & \textcolor{blue}{2.4 / 0.49 / 71.2}  \\
\hline
\end{tabular}
\label{indoor6_results_d2s_proposed}
\end{table}

\begin{table}
\centering
\caption{Ablation study on localization results when dropping components in the proposed architecture. SA represents self-attention.}
\begin{tabular}{c|c|c} 
\hline
\multirow{2}{*}{Settings} & Scene-1                               & Scene-2a                               \\
                          & (cm / deg. / \%)                      & (cm / deg. / \%)                       \\ 
\hline
No SA points \& lines                & 12.4 / 1.84 / 26.53                   & 7.70 / 0.79 / 26.85                    \\
No SA points              & 8.43 / 1.28 / 36.55                   & 5.80 / 0.56 / 45.53                    \\
No SA lines               & 4.44 / 0.74 / 56.45                   & 3.95 / 0.40 / 63.04                    \\
No pruning points \& lines  & 3.82 / 0.64 / 61.08                   & 3.50 / 0.37 / 66.54                    \\
No pruning points         & 3.79 / 0.63 / 62.20                   & 3.48 / 0.35 / 70.82                    \\
No pruning lines          & 3.69 / 0.60 / \textcolor{blue}{64.71}                   & \textcolor{blue}{3.11} / 0.35 / 72.37                    \\
\textbf{Attached full}    & \textcolor{blue}{3.62 / 0.59 }/ 63.95 & 3.13 / \textcolor{blue}{0.32 / 75.49}  \\
\hline
\end{tabular}
\label{ablation_dropping}
\end{table}

\begin{figure}
    \centering
    \includegraphics[width=\linewidth]{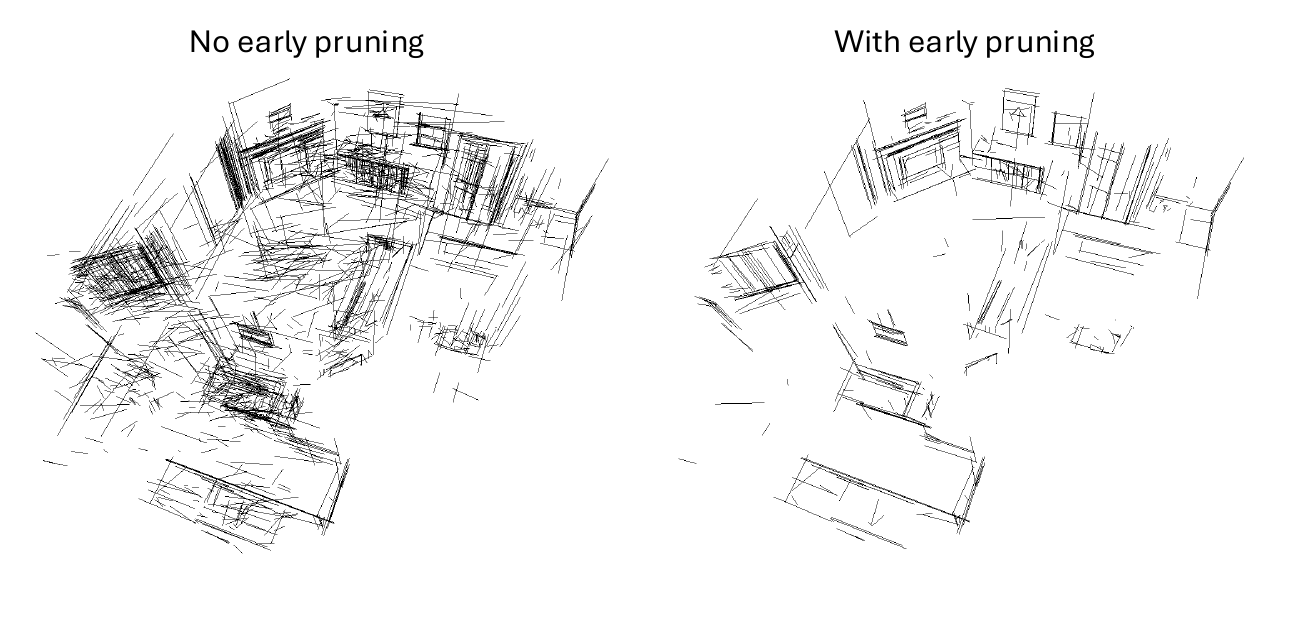}
    \caption{\textbf{Qualitative Results with Line Pruning Layer.} 
We present the qualitative results of predicted 3D lines from 20 randomly selected test images in Scene-1, using a stricter pruning threshold of \(\delta^l = 0.2\).}
    \label{pruning_lines3d}
\end{figure}

\begin{table}
\centering
\caption{Ablation study in increasing network params of PL2Map \cite{bui2024representing} in comparison with proposed method.}
\resizebox{\linewidth}{!}{%
\begin{tabular}{c|c|c|c} 
\hline
\multirow{2}{*}{Method} & \multirow{2}{*}{\# param.} & Scene-1                             & Scene-2a                               \\
                        &                            & (cm / deg. / \%)                    & (cm / deg. / \%)                       \\ 
\hline
PL2Map \cite{bui2024representing}                  & 6.4M                       & 4.73 / 0.84 / 51.7                  & 4.75 / 0.49 / 53.7                     \\
PL2Map* \cite{bui2024representing}               & 8.3M                       & 5.09 / 0.89 / 49.6                  & 3.66 / 0.41 / 60.7~                    \\
\textbf{Proposed}       & 8.3M                       & \textcolor{blue}{3.62 / 0.59/ 64.0} & \textcolor{blue}{3.13 / 0.32 / 75.5}  \\
\hline
\end{tabular}
}
\label{pl2map_increase_param}
\end{table}
In this section, we further evaluate the proposed methods on a more challenging dataset, Indoor-6 \cite{do2022learning}. Fig. \ref{indoor6_example} shows an example from the Indoor-6 dataset, which was captured under varying conditions between training and testing, posing significant challenges for regression-based methods. Indoor-6 consists of six different scenes, with images captured over multiple days, exhibiting substantial lighting variations. Each scene contains several thousand training and testing images.

\begin{table*}
\centering
\caption{Quantitative median pose errors and timing results on the Indoor-6 dataset when varying line detector methods during evaluation.  
The asterisk (*) indicates that the regressor was trained using high-quality lines detected by DeepLSD~\cite{pautrat2023deeplsd}.  
The forward time represents the average time required to generate 3D points and lines from an RGB image, including the point feature extractor, line detector, and two regression branches of the proposed method.  
The PnP time denotes the average time required to estimate camera poses from the predicted 3D lines and points.}
\resizebox{\linewidth}{!}{%
\begin{tabular}{c|c|c|c|c|c|c} 
\hline
\multirow{2}{*}{Indoor-6~~} & \multicolumn{2}{c|}{PL2Map \cite{bui2024representing} + DeepLSD \cite{pautrat2023deeplsd}}          & \multicolumn{2}{c|}{Proposed + DeepLSD \cite{pautrat2023deeplsd}}        & \multicolumn{2}{c}{Proposed* + LSD \cite{von2008lsd}}              \\ 
\cline{2-7}
                            & (cm / deg. / \%)  & (forward / PnP time / FPS) & (cm / deg. / \%)  & (forward / PnP time / FPS) & (cm / deg. / \%)  & (forward / PnP time / FPS)   \\ 
\hline
Scene-1                     & 4.7 / 0.84 / 51.7 & ~94.9 ms~/ 23.8 ms~/ 8.4   & 3.6 / 0.59 / 64.0 & 93.6 ms / 25.6 ms / 8.3    & 3.7 / 0.61 / 62.6 & 30.5 ms / 30.6 ms / 16.4     \\
Scene-2a                    & 4.8 / 0.49 / 53.7 & 99.5 ms~/ 11.2 ms~/ 9.0    & 3.1 / 0.32 / 75.5 & 98.8 ms / 17.0 ms / 8.6    & 3.3 / 0.36 / 74.3 & 31.1 ms / 17.0 ms / 20.8     \\
Scene-3                     & 5.3 / 1.05 / 49.2 & 99.3 ms~/ 44.4 ms~/ 7.0    & 3.1 / 0.62 / 65.8 & 97.1 ms / 56.0 ms / 6.5    & 3.3 / 0.63 / 64.1 & 34.2 ms / 67.3 ms / 9.8      \\
Scene-4a                    & 2.0 / 0.49 / 82.9 & 95.6 ms~/ 7.5 ms~/ 9.7     & 1.8 / 0.42 / 82.2 & 93.2 ms / 17.3ms / 9.0     & 1.6 / 0.42 / 82.2 & 33.7 ms / 19.4 ms / 18.8     \\
Scene-5                     & 7.7 / 1.21 / 40.3 & 95.8 ms~/ 15.0 ms~/ 9.0    & 6.0 /~0.87~/ 42.9 & 94.3 ms / 37.3 ms / 7.3    & 5.9 / 0.92 / 44.8 & 30.0 ms / 41.0 ms / 14.1     \\
Scene-6                     & 3.4 / 0.61 / 64.1 & 96.2 ms~/ 10.2 ms~/ 9.4    & 2.3 / 0.40~/ 74.9 & 94.6 ms / 18.8 ms / 8.8    & 2.3 / 0.40 / 72.1 & 31.1 ms / 20.0 ms / 19.6     \\ 
\hline
Average                     & 4.7 / 0.78 / 57.0 & 96.9 ms~/ 18.7 ms~/ 8.7    & 3.3 / 0.54 / 67.6 & 95.3 ms / 28.7 ms / 8.1  & 3.3 / 0.56 / 66.7 & 31.8 ms / 32.6 ms / 16.6  \\
\hline
\end{tabular}
}
\label{deeplsd2lsd}
\end{table*}

For comparison with previous re-localizers that integrate line features, we evaluated our method against the matching-based approach Limap \cite{liu20233d} and the regression-based method PL2Map \cite{bui2024representing}. However, there is a lack of regression-based methods that address both point and line features simultaneously. Therefore, we also compare our method with point-only regression methods, including DSAC$^{\star}$, ACE \cite{brachmann2023accelerated}, NBE+SLD \cite{do2022learning}, and D2S \cite{bui2024d2s}. The results for DSAC$^{\star}$ \cite{brachmann2021visual} and NBE+SLD \cite{do2022learning} on this dataset are taken from \cite{do2022learning}, while the results for ACE \cite{brachmann2023accelerated} and D2S \cite{bui2024d2s} were evaluated by \cite{bui2024d2s}. Since PL2Map \cite{bui2024representing} and Limap \cite{liu20233d} have not been previously evaluated on Indoor-6, we used the same configurations provided by the authors for evaluation on this dataset.

We present the results of the proposed method in comparison with previous works in Table \ref{indoor6_results} and qualitative results are shown in Fig. \ref{qualitative_result_indoor6}.  The findings indicate that the proposed method, which separately regresses points and lines, achieves the best re-localization accuracy among other regression-based methods. Notably, the proposed method shows significant improvements over the previous line-assisted method, PL2Map \cite{bui2024representing}, with accuracy improvement gains of approximately 12.3\%, 21.8\%, 16.6\%, and 10.8\% in Scene-1, Scene-2a, Scene-3, and Scene-6, respectively. A slight improvement of about 2\% is also observed in Scene-4a and Scene-5. Additionally, the proposed method narrows the performance gap with FM-based methods such as HLoc \cite{sarlin2019coarse, sarlin2020superglue} and Limap \cite{liu20233d}, particularly in challenging, changing conditions. In Fig. \ref{qualitative_result_indoor6}, we show the qualitative results of predicted 3D line maps and estimated camera poses by the proposed method. 

Overall, our method significantly improves point-line mapping regression, surpassing previous state-of-the-art regression-based methods on the Indoor-6 dataset.

\subsection{Ablation Study}
In this section, we present additional experiments with the proposed method to provide a more detailed analysis.

\textbf{Localization with point branch only.} Since the proposed method can localize using only predicted 3D points, we compare it with the point-based method ACE \cite{brachmann2023accelerated} in Table \ref{indoor6_results_d2s_proposed}. Interestingly, the proposed method, when using points alone, significantly outperforms ACE on this dataset, demonstrating its superior adaptability to changing conditions.

\textbf{Architecture analysis.} To justify the design choices of the proposed method, we present ablation results in Table \ref{ablation_dropping}, where various components of the architecture are systematically removed. Additionally, Fig. \ref{pruning_lines3d} provides a qualitative comparison of predicted 3D lines with and without pruning layers, including an example where line pruning is applied with $\delta^{l} = 0.2$. Without pruning layers, the model generates numerous incorrect and redundant 3D lines, which can negatively impact camera pose estimation. These results confirm that each component is essential for achieving the highest performance in the proposed architecture.

\textbf{Comparison with PL2Map.} Since the proposed network has a larger number of parameters compared to PL2Map \cite{bui2024representing}, we re-ran PL2Map with an increased number of parameters by adding more self and cross attention layers to match the proposed method. These results are reported in Table \ref{pl2map_increase_param} in comparison with our method, where the asterisk (*) indicates this PL2Map setting.

\textbf{System Efficiency}. The proposed method uses DeepLSD \cite{pautrat2023deeplsd} by default for 2D line detection and labeling via Limap \cite{liu20233d}, as it yields higher-quality line segments compared to earlier approaches, particularly LSD \cite{von2008lsd}. However, DeepLSD can be computationally expensive for certain downstream applications. Therefore, we investigated replacing DeepLSD with LSD \cite{von2008lsd} during the testing phase to improve computational efficiency. Interestingly, the method continues to perform well on LSD-detected lines, despite LSD producing more discontinuous or spurious segments.

As reported in Table \ref{deeplsd2lsd}, the re-localization error increases only slightly when using LSD segments, while the average FPS improves significantly from 8.1 to 16.6. Moreover, despite employing a larger network architecture, the proposed method achieves a comparable inference speed to PL2Map \cite{bui2024representing}, with an FPS of 8.7 versus 8.1, while demonstrating a significant accuracy improvement of 18.6\%.

\section{Conclusions}

We proposed a new approach designed to improve 3D points and line regression from 3D SfM models of points and lines. Our method improves camera localization accuracy by focusing the network's attention on key features during training. This is achieved through a new seperative architecture, followed by self-attention mechanisms that prioritize the most relevant features. Additionally, the method benefits from early pruning of non-robust descriptors, allowing the network to focus more on the most robust descriptors. Experimental results show that our approach achieves a relative improvement gain of up to 21.8\% on the Indoor-6 dataset compared to the previous point-line regression method.



\bibliographystyle{IEEEtran}
\bibliography{reference.bib}

\end{document}